\documentclass{article}
\usepackage{spconf,amsmath,graphicx}
\usepackage{amssymb} 

\usepackage{booktabs}       
\usepackage{multirow, makecell}
\usepackage[hyphens]{url}  
\usepackage{xcolor}
\usepackage{pifont}
\newcommand{\cmark}{\ding{51}}%
\newcommand{\xmark}{\ding{55}}%
\usepackage{CJKutf8}
\usepackage{hyperref}
\usepackage{enumitem}

\title{Neural Inverse Text Normalization}
\name{Monica Sunkara, Chaitanya Shivade, Sravan Bodapati, Katrin Kirchhoff}
\address{Amazon AWS AI}
\begin{document}
\ninept
\maketitle

\begin{abstract}
While there have been several contributions exploring state of the art techniques for text normalization, the problem of inverse text normalization (ITN) remains relatively unexplored. The best known approaches leverage finite state transducer (FST) based models which rely on manually curated rules and are hence not scalable. 
We propose an efficient and robust neural solution for ITN leveraging transformer based seq2seq models and FST-based text normalization techniques for data preparation.
We show that this can be easily extended to other languages without the need for a linguistic expert to manually curate them. We then present a hybrid framework for integrating Neural ITN with an FST to overcome common recoverable errors in production environments.
Our empirical evaluations show that the proposed solution minimizes incorrect perturbations (insertions, deletions and substitutions) to ASR output and maintains high quality even on out of domain data. 
A transformer based model infused with pretraining consistently achieves a lower WER across several datasets and is able to outperform baselines on English, Spanish, German and Italian datasets.


\end{abstract}


\section{Introduction}
Inverse Text Normalization (ITN) is the process of converting spoken form of output from an automatic speech recognition (ASR) system to the corresponding written form. The written form is more commonly consumed by users and automated downstream processes. This includes simple entities such as cardinals, ordinals, and currencies, as well as complex entities involving addresses, email id, dates, times, and URIs. ITN is closely related to the task of text normalization where the goal is to convert text from written form to the spoken form \cite{allen1987text}.

Exploiting the power of neural networks for text normalization has been challenging until recently when recurrent neural networks (RNNs) were the state of the art \cite{sproat2016rnn}. While RNNs produced great results when evaluated using standard metrics, their errors made them infeasible for adoption in practical systems. More recently,  \cite{pramanik2019text} successfully employed a memory augmented neural network architecture that is independent of FSTs for text normalization. Similar success has been noted where neural generation models replaced rule based systems for tasks such as ASR spelling correction \cite{guo2019spelling} and grammar correction \cite{hrinchuk2020correction, choe2019neural}.

 Although FST based approaches to ITN work well \cite{mansfield2019neural}, they are expensive to scale across languages since native language speakers are needed to curate transformation rules. We therefore investigate applicability of neural network based models for ITN and evaluate them using WER specifically for ITN and non-ITN parts of the text. While RNNs are powerful for sequence to sequence tasks, transformer based models \cite{devlin-etal-2019-bert} offer pretraining abilities using vast amounts of data. However, incorporating pretrained models is not trivial and is often specific to the task \cite{zhu2020incorporating}.

In light of the gaps with existing methods and resources for ITN, we present results with a neural approach towards the task. Specifically, we make the following contributions:
\begin{itemize}[leftmargin=*,itemsep=0pt, topsep=0pt]
    \item Since data required for training ITN models is not publicly available and is hard to collect, we employ a data generation pipeline for ITN using a text to speech (TTS) frontend. 
    \item We propose a robust neural ITN solution based on seq2seq models that can perform well across domains and noisy settings such as conversational data.
    \item We explore different ways to leverage pretrained models for ITN and demonstrate their superior performance across several domains and on ASR transcribed data.
    \item Finally, we show the first results for three of the European languages in mono-lingual and multi-lingual settings.
\end{itemize}

\vspace{-2mm}
\section{Text processing pipeline}
\label{sec:text_processing_pipeline}

In conventional approaches, ITN is addressed by using a set of rules through FSTs. However, neural models can be data hungry and may require good amount of labelled data for supervised learning. Collecting large amount of annotations at scale is prohibitive in terms of cost, resources, and time. Therefore, we employ FST-based text normalization methods to automatically normalize written form of text. This is similar to synthetic data generation employed successfully in the past \cite{peyser2019improving}. However, the data prepared in such a way poses a number of problems for modeling ITN:

\begin{itemize}[leftmargin=*,itemsep=0pt, topsep=0pt]
    \item Table \ref{tab:spoken_form} shows examples of spoken-form input and written-form output.
    As shown in the first example, there can be several variations of a written form of text, but text normalization techniques always use one fixed variation. This problem is further aggravated with variations in locale, and language.
    \item Text normalization systems often ignore several punctuation symbols since the goal is to produce spoken form of text. However, punctuation is often relevant in the written form. For example, if the TTS system omits quotation marks during the conversion, it is important to restore them in spoken form output for efficient modeling of ITN.
    \item Text normalization techniques may introduce errors when normalizing numbers or expanding some short forms, as the conversion requires to disambiguate based on the context. e.g., Lakeside Dr. is normalized as Lakeside drive where as Dr. John is converted as Doctor John. 
\end{itemize}

\begin{table}[t]
\begin{tabular}{ll}
\toprule
Spoken form & Written form  \\ \midrule
\begin{tabular}[c]{@{}l@{}}Two thousand one hundred five\\ Two thousand one hundred and five\\ Twenty one oh five\\ Two one zero five\\ Two one oh five\end{tabular} & 2105 \\ \midrule 
october twenty twenty twenty & October 20, 2020 \\
four percent of five dollars is twenty cents & 4\% of \$5 is 20 cents \\
\bottomrule
\end{tabular}
\caption{Examples of spoken-form input and written-form output}
\label{tab:spoken_form}
\end{table}

To address first issue, we create synthetic data by randomly sampling from cardinal numbers related sentences and introducing additional variations required for modeling ITN in spoken-form of text. This synthetic data is augmented to the text data created using text normalization techniques for experiments presented in Section \ref{sec:results}.
We used Levenshtein edit-distance to align written form with spoken form in order to restore punctuation marks.

\section{Models}
We treat the traditional FST model as a baseline and discuss architectures for neural models. Finally, we describe a hybrid model for production settings.

\subsection{Finite State Transducer}
Our conventional baseline approach is a Finite State Transducer (FST) constructed using JFLAP \footnote{http://www.jflap.org/}. Each state in the FST performs a series of edits to the input string to get its corresponding written format output string. Our FST model covers a wide range of entities which do not require contextual understanding or disambiguation such as: Cardinals, Fractions, Ordinals, Years, floats, Date, Time, Currency, Units and Measurements.

\subsection{Neural ITN}
\label{sec:rnn}
We model inverse text normalization as a sequence-to-sequence problem where the source and target are spoken and written form of text respectively. Our seq-to-seq baseline model is similar to \cite{mansfield2019neural} and uses Bahdanau content-based additive attention \cite{bahdanau2014neural} to align the output sequence with input. We also implement a non-recurrent transformer based sequence to sequence model based on Vasvani et al., \cite{vaswani2017attention}, where multi-head self-attention layers are used in encoder and decoder. For all our transformer models, the source and target sentences are segmented into sub word sequences. 

\vspace{2mm}
\noindent \textbf{Copy Attention:} For a task like ITN, there exists a significant overlap between source and target sentences. Standard sequence-to-sequence models which rely on content based attention are often prone to undesirable errors such as insertions, substitutions and deletions during conversion. In order to reproduce the target close to the source sentence, we adapt the transformer based sequence-to-sequence models by utilizing a copying mechanism as described in \cite{see2017get}. The copy mechanism uses a generation probability to choose between source sentence vocabulary and a fixed target vocabulary thus allowing to generate out-of-vocabulary (OOV) words. The generation probability for each time-step is computed from context vector, transformer decoder self-attention output and the decoder input. 

\vspace{2mm}
\noindent \textbf{Pretrained models for ITN:} Recently, pre-trained models such as BERT \cite{devlin2019bert} and ELMo \cite{peters2018deep} have achieved tremendous success and lead to significant improvements on various natural language understanding tasks. These models trained on large amounts of unlabelled data capture rich contextual representations of the input sentence. In this work, we attempt two strategies to incorporate pretraining into ITN: (1) To use a pre-trained sequence-to-sequence model and fine-tune it for ITN, and (2) Using a pre-trained masked language model like BERT as context-aware embeddings for ITN. For the first strategy, following \cite{lewis2019bart}, we initialize the encoder and decoder of a sequence-to-sequence model with a pre-trained BART model, and then fine-tune the model on ITN datasets. Unfortunately, we did not observe good performance. For the second strategy, following the practice of \cite{zhu2020incorporating}, we use BERT to extract context-aware embeddings and fuse it into each layer of transformer encoder and decoder via an attention mechanism.

\subsection{Hybrid solution}
\label{sec:hybrid}
Despite incorporating the mechanisms we discussed in previous section, sequence-to-sequence models are still prone to some errors. Also, in a real world application, neural models require a good number of training examples to handle each new entity. In order to overcome these issues, we propose a novel hybrid approach combining neural ITN with an FST, where the spoken form output of ASR system is first passed through the proposed neural ITN model, followed by an FST. A confidence score emitted by the neural ITN model is used as a switch to make a decision in the run time whether to use neural ITN output. 
This solution has a three-fold advantage: (1) Having neural ITN as the first pass, helps to normalize the input sentence well based on a deeper contextual understanding, (2) With FST as the final component, we can correct any common neural ITN errors by adding rules to FST, and (c) In the case of any unrecoverable mistakes by ITN in the run time, the confidence score can be used as a safety measure from outputting erroneous neural ITN output.

\section{Experimental setup}

\subsection{Data}
All our experiments were conducted on data extracted from publicly available Machine Translation, Speech Translation, and Text Summarization data sources. For monolingual English models we used Wikipedia data subsampled from the publicly available release of parallel written-spoken formatted text from \cite{sproat2016rnn}. 
We also test our models on additional domains such as crawl, ted talk, and news. For this purpose, we extracted data from ParaCrawl, NewsCommentary (News-C) data sets provided by Shared Task: Machine Translation of News\footnote{http://www.statmt.org/wmt20/translation-task.html} and MuST-C dataset release from \cite{di2019must} and CNN-DailyMail data sets from DeepMind Q\&A Dataset\footnote{https://cs.nyu.edu/~kcho/DMQA/}. 
For multilingual experiments, we used the data extracted from ParaCrawl, NewsCommentary (News-C) and MuST-C \cite{di2019must} for German, Spanish and Italian languages. 

We prepare parallel written-spoken form for our datasets in the following manner. We extract the text in written form from public sources and generate the corresponding spoken form using FST based text normalization of a commercial text to speech service\footnote{https://aws.amazon.com/polly}.
Finally, we augment the English data with synthetically generated data as described in section \ref{sec:text_processing_pipeline}.
The English model data splits contain 600k, 130k sentences for training and tuning respectively. The Multilingual models train and tune splits contain 200k, 25k sentences respectively. Apart from the above mentioned datasets, as a check for robustness, we also evaluate our monolingual English models on ASR outputs (500 sentences each for broadcast and conversational domains) generated from private data sets. The total number of ITN entities is 5\% of words in the test data.

\subsection{Metrics}
Previous work on inverse text normalization and related tasks measures the performance of models by computing word error rate (WER) on the entire target sentence. 
In our work, we introduce two additional metrics namely ITN WER (I-WER) and Non-ITN WER (NI-WER) focusing on the quality of text normalization and to measure undesirable perturbations to original source text respectively.
The metrics are computed as follows: First, each word in the written form reference in tagged with ITN or N-ITN indicating if a word should be normalized or should be copied to target respectively. In order to perform the tagging, we align the source (spoken) and reference (written) sentences using levenshtein distance. The words which are common in both sentences are tagged as N-ITN and the rest as ITN in the reference. ITN WER (I-WER) is computed by summing up the substitutions, insertions, and deletions between the target hypothesis (ITN output) and reference and dividing it with number of reference words considering only those tagged as ITN. Non-ITN WER (NI-WER) is computed similarly over only N-ITN tagged words. The overall WER is computed across all the reference words.

\begin{table}[t]
\begin{tabular}{lcccc}
\toprule
Model        &  Syn & WER & I-WER & NI-WER   \\ \midrule
FST       & \xmark & 14.4    & 103.8 & 0.2 \\
RNN-Word & \xmark & 12.9 & 19.8 &	0.9 \\
RNN-Subword    & \xmark & 2.2 &  14 & 0.4 \\ \midrule
Transformer  & \xmark & 1.3 & 6.2 & 0.6 \\
 & \cmark & 1.1 & 5.2 & 0.4  \\ \midrule
Transformer &  &  &  &  \\
\hspace{0.5em}+ Copy Attention   & \cmark & 1.0	 & 5.2	& 0.3  \\
\hspace{0.5em}+ BERT-fusion & \cmark & 0.9  & 4.8 & 0.3 \\
BART                        & \cmark & 14.5 & 72.7 & 5.2 \\
 \bottomrule
\end{tabular}
\caption{Comparison of models on Wikipedia test set.}
\label{tab:main_results}
\end{table}

\subsection{Model Configurations}
As sequence-to-sequence baselines, we implement a word and subword based recurrent neural network models similar to \cite{mansfield2019neural}. The RNN model consists of a 2-layer bi-directional long short term memory network (bi-LSTM) encoder and a 2-layer LSTM decoder, with 512 hidden states for the bi-LSTM. Decoding uses the attention mechanism from \cite{bahdanau2014neural}. The second sequence to sequence architecture is a subword transformer model \cite{vaswani2017attention} with 12 layers. We also implement two additional mechanisms in to our subword transformer model. The first mechanism is copy attention mechanism similar to \cite{see2017get}. The second mechanism is BERT-Fusion into the transformer based encoder and decoder layers following \cite{zhu2020incorporating}. For English experiments we use bert-base-uncased to extract pretrained representations for an input sequence, whereas for multilingual experiments we use bert-base-multilingual-uncased\footnote{https://github.com/bert-nmt/bert-nmt/blob/master/bert/modeling.py}.
In our experiments, for a given sentence, subwords are generated through a model trained using the SentencePiece toolkit\footnote{https://github.com/google/sentencepiece}. We use a subword inventory size of 16k\footnote{In our ablation study experiments conducted in the range of 2k-32k subwords, 16k gave best results on Wikipedia} and a beam size of 5 for all our experiments.

\begin{table}[t]
\begin{tabular}{lclccc}
\toprule
Dataset   & ASR & Model  & WER  & I-WER & NI-WER \\ \midrule
CNN       & \xmark    & Copy Att & 2.0 & 29.7 & 0.4 \\
           &   & BERT-fusion & 2.2 & 19.5 & 0.6 \\ 
DailyMail  & \xmark  & Copy Att & 2.4 & 17.8 & 0.4 \\
           &   & BERT-fusion & 1.7 & 3.9 & 0.4 \\ 
News-C & \xmark  & Copy Att & 1.5 & 19.4 & 0.2 \\
        &      & BERT-fusion & 1.2 & 10.4 & 0.3 \\ 
MuST-C  & \xmark      & Copy Att & 2.7 & 23.2 & 1.4 \\
         &     & BERT-fusion & 3.3 & 20.4 & 1.7 \\ 
Paracrawl  & \xmark   & Copy Att & 7.9 & 29.2 & 0.8 \\
           &   & BERT-fusion & 3.9 & 1.8 & 3.9 \\ \midrule
BCN  & \cmark   & Copy Att & 3.3 & 53.3 & 0.7 \\
           &   & BERT-fusion & 3.2 & 44.5 & 1.1 \\ 
TEL  & \cmark   & Copy Att & 5.3 & 65.4 & 1.5 \\
           &   & BERT-fusion & 4.3 & 46.9 & 1.6 \\
\bottomrule
\end{tabular}
\caption{Performance across various domains including News, Ted Talks, Crawl etc.}
\vspace{-4mm}
\label{tab:domaindatasets}
\vspace{-4mm}
\end{table}

\section{Results}
\label{sec:results}
Table \ref{tab:main_results} summarizes our experiments on Wikipedia dataset. We report an overall word error rate (WER) at sentence level, ITN WER (I-WER) and Non ITN WER (NI-WER). First, we compare the performance of the baseline Finite State Transducer (FST) model, word-based RNN, subword RNN and Transformer models. The RNN subword model performs better than the word based RNN and FST, with a significant relative improvement of 84.7\% in terms of WER when compared to FST model. We must state that the FST model does not cover all normalization entities in the Wikipedia data leading to a higher I-WER. The transformer model outperforms others with a WER of 1.3. In addition, synthetic data augmentation leads to an improvement in both ITN and non-ITN WER and there by a relative overall WER improvement of ~18\%. We then present the experiments on transformer model by incorporating additional mechanisms such as copy attention and BERT-fusion for further improvement. Copy attention leads to a relative improvement of 25\% in non-ITN WER. This is expected, as copy mechanism helps in reducing the out of vocabulary (OOV) problem by copying the words from source to text and enforces the target sentence to remain close to the source. Transformer model fused with pretrained BERT representations shows improvement in both ITN and non-ITN WER performing the best of all models. Since the BERT model is pretrained on large amounts of written text from public sources such as Wikipedia, fusing its representations into transformer model seems to result in gains than models trained from scratch. Also, since we trained and tested on the same domain data, a transformer with BERT-fusion performs comparably well even without copy mechanism. This motivated us to extend our experiments and test the performance of neural ITN models on several other domains as illustrated in section \ref{sec:real}. 

We also validate using a pre-trained sequence-to-sequence model and fine-tuning it for ITN task would help. For this purpose, we finetuned a pretrained BART model on ITN data and evaluated its performance on Wikipedia dataset. Unfortunately, we did not observe good performance by using the BART model out of the box. The decoder outputs had several insertions and failed to normalize for even simple entities. 

\begin{table}[t]
\begin{tabular}{lcccccc}
\toprule    
\multirow{2}{*}{Usecase}   & \multicolumn{3}{c}{BCN}  & \multicolumn{3}{c}{TEL} \\ \cline{2-7}
 & FST & Neural & Hybrid & FST & Neural & Hybrid \\ \midrule
Numbers & 2.8 & 2.6 & 2.5 & 5 & 4.5 & 4.2 \\
Units & 7.5 & 5.6 & 5.4 & 4.4 & 3.6 & 3.4 \\
Date Time & 5 & 1.8 & 1.6 & 7 & 6 & 5.8 \\
Misc & 10 &  3.5  &  3.2 & 15.6 & 11.1 & 11.1 \\ 
\bottomrule               
\end{tabular}
\vspace{-4mm}
\caption{Performance on ASR outputs (Broadcast and Conversational) in terms of WER (\%).}
\vspace{-4mm}
\label{tab:asr}
\end{table}

\vspace{-2mm}
\subsection{Towards a real world system}
\label{sec:real}

We carried out evaluations of our best neural ITN models namely copy attention and BERT-fusion across several domains. Results presented in Table \ref{tab:domaindatasets} show that transformer model with copy attention (Copy Att) consistently outperformed BERT-fusion model w.r.t Non ITN WER (NI-WER) across all the domain testsets. Copy attention seems to play an effective role in copying the source sentence words into the target for out of domain datasets and OOV words, whereas using pretrained embeddings seem to be of limited help. On the other hand, w.r.t ITN WER (I-WER) BERT-fusion significantly outperformed copy attention across all test sets. This explains that, fusion with representations extracted from a model pretrained on large written datasets helps in improving the normalization. Moving towards a real world system, we evaluate on ASR outputs from broadcast (BCN) such as news, media etc and conversational (TEL) domains. BERT-fusion performs better than copy attention on both the datasets due to significant improvements in ITN WER. These observations motivated us to further explore their performance across different entities. We compare entity wise performance of FST, best neural ITN model (BERT-fusion) and hybrid solution described in section \ref{sec:hybrid}. Though neural ITN model significantly outperforms FST (see Table \ref{tab:asr}) across all entities, the model outputs often contain some common errors. Also, the higher WER for FST in Misc category shows the necessity to define manually curated rules for each entity. However, using a hybrid approach, we observe that few of the recoverable common errors can be corrected, there by leading to an additional improvement in performance. We also present some qualitative examples from our best model (BERT-fusion) in Table \ref{tab:examples}. 

\begin{table}[h]
\begin{tabular}{p{8cm}}
\toprule
\textbf{Input:} contact number for us is \emph{ one eight hundred two five five seven eight two eight} \\
\textbf{Output:} contact number for us is \emph{1-800-255-7828} \\
\midrule
\textbf{Input:} any time not not at \emph{noon} or \emph{two}  or \emph{four} \\
\textbf{Output:} any time not not at \emph{12:00 pm} or \emph{2:00} or \emph{4:00} \\
\midrule
\textbf{Input:} it was priced at \emph{three thousand six four nine point eight four dollars}  \\
\textbf{Output:} it was priced at \emph{\$3649.84} \\
\textbf{FST:} it was priced at  \emph{3006 4 \$9.84}  \\
\midrule
\textbf{Input:} \emph{ten twenty nine}  gmt november \emph{twenty ninth twenty   twelve} \\
\textbf{Output:} 10:29 gmt november 29       2012 \\ 
\midrule
\textbf{Input:} we think certainly \emph{ten to fifteen thousand dollars}      a month
  \\
\textbf{Output:} we think certainly \textcolor{red}{\emph{\$10-\$15000}}    a month \\
\midrule
\textbf{Input:} \emph{florida three three nine six oh}    yusa or phone \emph{nine four one four six five four three two one}          fax \emph{four six}
  \\
\textbf{Output:} \emph{fl 33960} yusa or phone \emph{941-465-4321} fax \emph{46} \\
\bottomrule
\end{tabular}
\vspace{-2mm}
\caption{Example utterances from Neural ITN outputs}
\label{tab:examples}
\vspace{-4mm}
\end{table}

\subsection{Performance of Multilingual Neural ITN models}
We extend our experiments to other European languages such as German, Spanish and Italian. For each language, we train two models: a vanilla transformer and a transformer model with BERT-fusion. In addition, we also train a multilingual model on data combined from all three languages. During training and inference, we add an artificial language token prefix to each sentence in both source (spoken-form) and target (written-form) similar to \cite{johnson2017google}.

The results are presented in Table \ref{tab:multilingual}. From the results, we observe that BERT-fusion consistently outperformed vanilla Transformer and the multi-lingual model outperformed the models trained on data from each individual language. It is clear that the multilingual model is greatly benefited from having lexical similarity among all three languages trained. Also, from the qualitative analysis of the results, we infer that  the coverage of subword units for each monolingual model is vastly improved in multilingual model by combining data from all three languages, thus leading to fewer number of OOVs.

\begin{table}[h]
\centering
\begin{tabular}{lccc}
\toprule
Model                             & German & Spanish & Italian \\ 
\midrule
Transformer                       &   3.5    &  6.1    &  2.3  \\ 
\hspace{0.5em}+ BERT-fusion       &   2.7   &   5.2  &   2.2   \\ 
\hspace{1em}+ Multi-lingual       &   2.1    &  5 &   1.6    \\ 
\bottomrule
\end{tabular}
\caption{Performance of neural ITN on multiple languages in terms of WER (\%)}
\vspace{-5mm}
\label{tab:multilingual}
\end{table}

\section{Related work}
\label{sec:related_work}

While ITN as a problem is not new \cite{shugrina2010}, work addressing the task using end to end neural sequence generation has been very limited. Most recently, \cite{ihori2020}
used pointer generator networks for ITN in Japanese. They limit their evaluation to a single dataset and note that the copy mechanism from the architecture \cite{see2017get} is key to performance improvements. ITN is one of the many tasks addressed in the larger problem of ASR post-processing for readability (APR) proposed in \cite{liao2020improving}. They compare several transformer based models across different openly available datasets and show that RoBERTa \cite{liu2019roberta} is a better model in most scenarios. While this is the closest study to our work, their evaluation is for the APR task which has a wider scope but is limited to monolingual English. In \cite{Pusateri2017AMD}, a three step approach is proposed for modeling ITN as a labeling problem, first identifying the candidates for transformation, second mapping the spoken candidates to written form, and finally applying post-processing if necessary. They infuse machine learning into rule based models by adopting LSTMs for the first step and using FSTs for the second.

\section{Conclusion}
We introduced a neural ITN approach leveraging pretrained models that can perform well across various domains covering more than 30 entities. Our results suggest that the proposed neural ITN solution can significantly outperform traditional FST based solutions on ASR outputs of unstructured datasets like Conversational domain. We have presented entity wise performance analysis and a hybrid solution for combining FST with neural ITN to minimize any common recoverable errors made by pure neural models. We showed that our solution can be easily extended to other languages by presenting the first multilingual ITN results, without the need for a linguistic expert. Future work will focus on combining copy attention with pretrained models to reap benefits from both the approaches.


\bibliographystyle{IEEEbib}
\bibliography{strings,refs}

\end{document}